\documentclass[letterpaper, 10 pt, conference]{ieeeconf}  

\IEEEoverridecommandlockouts                              

\usepackage{amsmath}
\usepackage{amssymb}
\usepackage{graphicx}
\usepackage{balance}
\usepackage[hidelinks]{hyperref}
\usepackage{xurl}
\usepackage{array}

\usepackage{multirow}
\usepackage{cite}

\usepackage[linesnumbered,ruled,vlined]{algorithm2e}

\title{\LARGE \bf
GenerativeMPC: VLM-RAG-Guided Whole-Body MPC with Virtual Impedance for Bimanual Mobile Manipulation
}

\author{Marcelino Julio Fernando, Miguel Altamirano Cabrera, Jeffrin Sam,\\
Yara Mahmoud, Konstantin Gubernatorov, and Dzmitry Tsetserukou%
\thanks{The authors are with the Intelligent Space Robotics Laboratory,\protect\\
Skolkovo Institute of Science and Technology, Moscow, Russia.\protect\\
{\tt\small \{Marcelino.Fernando,M.Altamirano,\protect\\%
Jeffrin.Sam,Yara.Mahmoud,\protect\\%
Konstantin.Gubernatorov,D.Tsetserukou\}\protect\\%
@skoltech.ru}}%
}

\begin{document}
\bstctlcite{IEEEexample:BSTcontrol}

\maketitle
\thispagestyle{empty}
\pagestyle{empty}

\begin{abstract}


Bimanual mobile manipulation requires tight integration between high-level semantic reasoning and safe, compliant physical interaction. End-to-end models often encode this relationship opaquely, whereas classical controllers lack semantic context. This paper presents GenerativeMPC, a hierarchical cyber-physical framework that maps visual and linguistic context to interpretable control parameters for a bimanual mobile manipulator. A Vision-Language Model with Retrieval-Augmented Generation (VLM-RAG) selects velocity limits and a safety margin for a reduced task-space Whole-Body Model Predictive Controller (MPC), while scheduling virtual stiffness and damping gains for an impedance-admittance controller, enabling context-aware compliance during human-robot interaction. An experience-driven vector database retrieves previously validated parameter sets without model retraining. Human-aware navigation is evaluated in simulation and on a differential-drive platform, where the commanded velocity is reduced by 60\% near a detected person while maintaining the selected experimental separation threshold. Bimanual manipulation is evaluated in simulation, achieving an 85\% overall task-success rate and final end-effector errors below 5 mm, with the corresponding hardware gains proposed for future deployment. 


\end{abstract}

\section{Introduction}\label{sec:intro}

\begin{figure}[!t]
    \centering
    \includegraphics[width=\columnwidth]{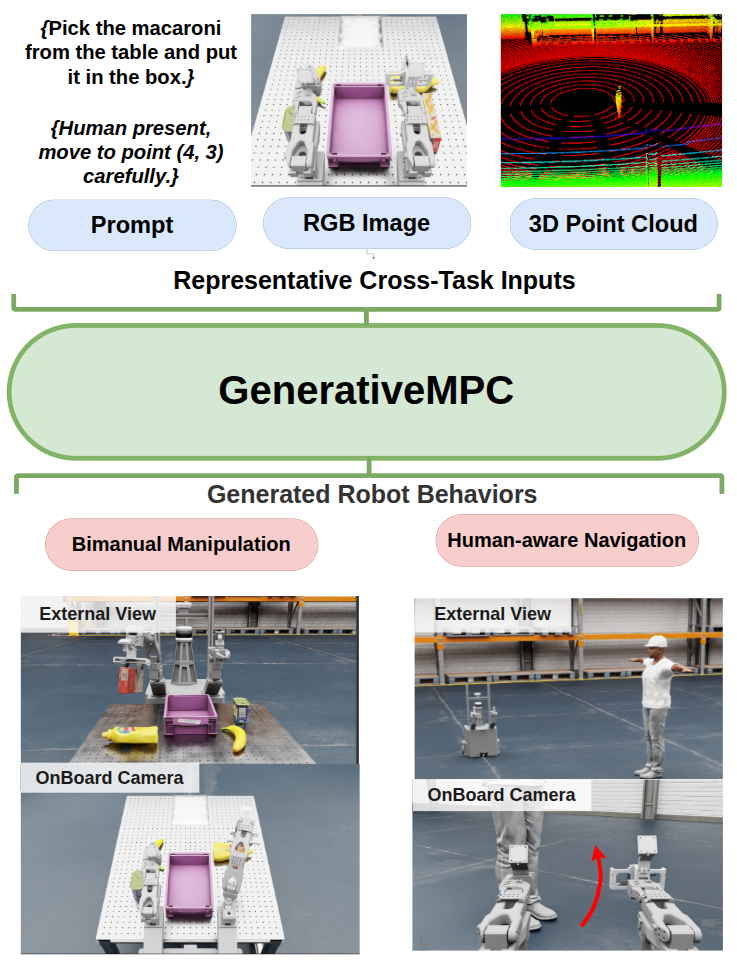}
    \caption{
    GenerativeMPC maps language instruction, RGB observation, and LiDAR point cloud to context-dependent control parameters for bimanual manipulation and human-aware navigation. Manipulation is shown in simulation and navigation on the physical platform, each from external and onboard views.
    }
    \label{fig:teaser}
    \vspace{- 0.5 cm} 
\end{figure}

Autonomous robots operating in human-centered environments must navigate unstructured spaces, manipulate objects with varying compliance requirements, and interpret instructions from non-expert users. Bimanual mobile manipulators, platforms combining a mobile base with two articulated arms, are uniquely positioned to address these demands. However, their complexity exposes a fundamental tension: the semantic richness required for task understanding and the physical precision required for safe interaction have been handled by separate, incompatible systems. This fragmentation prevents a unified loop where high-level reasoning directly modulates the robot's physical response.

\begin{figure*}[t!]
    \centering
    \includegraphics[width=0.85\textwidth]{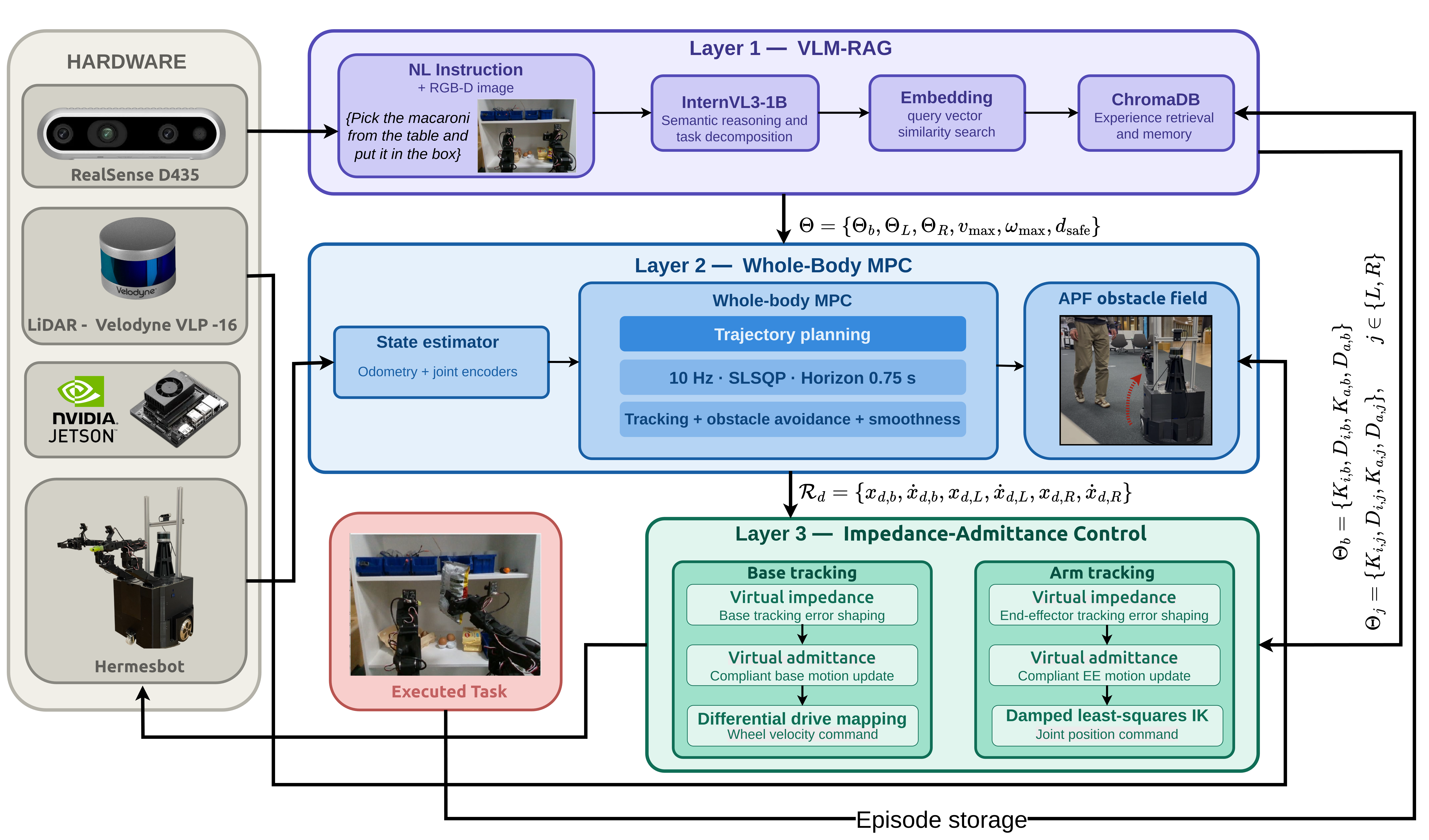}
    \vspace{- 0.2 cm} 

    \caption{GenerativeMPC three-layer architecture. Layer~1 (VLM-RAG) supplies velocity limits and a safety margin $d_{\text{safe}}$ to the Whole-Body MPC, and stiffness and damping gains $K_i$ and $D_i$ to the impedance-admittance controller. Layer~2 plans trajectories at 10\,Hz over $N=5$ steps with embedded APF avoidance; Layer~3 executes compliant tracking at 50\,Hz for the base and both arms.}
    \label{fig:architecture}
    \vspace{- 0.6 cm} 
\end{figure*}

Recent advances in Vision-Language Models (VLMs) have demonstrated remarkable zero-shot reasoning over visual scenes~\cite{c1, c2}. However, existing VLM-based planners typically output discrete subgoals, leaving physical compliance, how stiff or yielding the robot should be on contact, undefined at the semantic level~\cite{c8, c10}. Recent models such as MoManipVLA~\cite{c16} and FALCON~\cite{c17} have advanced end-to-end mobile manipulation, while contact-rich policies such as HapticVLA~\cite{hapticvla} learn compliant interaction directly from data. These systems pursue kinematic or task success while keeping the robot's physical ``feel'' (stiffness and damping) implicit in learned weights, rather than exposing it as an interpretable, context-conditioned parameter set. Separately, Whole-Body Model Predictive Control (MPC) frameworks enable real-time trajectory optimization for high-degree-of-freedom platforms~\cite{c4, c5}. These methods have achieved impressive results in collision-free locomotion and manipulation~\cite{c6}, while other approaches combine RL-based planning with low-level MPC~\cite{c7}. However, the velocity constraints and safety margins that shape their behavior are typically fixed at design time.

The present work addresses the open problem of grounding semantic scene understanding into predictable physical control for bimanual mobile manipulation. Building on the semantic-to-compliance paradigm established for fixed-base, single-arm systems in SafeHumanoid~\cite{c14} and HumanoidVLM~\cite{humanoidvlm}, we extend this logic to a unified cyber-physical framework where Whole-Body coordination, nonholonomic navigation, and dual-arm compliance are handled jointly.

Furthermore, we address the robotics ``memory gap'': Retrieval-Augmented Generation (RAG) grounds language outputs in factual context, but its use in experience-driven control loops remains largely unexplored.

This paper presents \textit{GenerativeMPC} (Fig.~\ref{fig:teaser}), a three-layer framework in which a VLM-RAG module, comprising InternVL3-1B~\cite{c_internvl} and ChromaDB, produces, from the camera image and task instruction, two sets of parameters: (1) velocity limits $v_{\max}$, $\omega_{\max}$ and safety margin $d_{\text{safe}}$ that constrain the Whole-Body MPC, and (2) stiffness $K_i$ and damping $D_i$ gains for a unified impedance--admittance controller. The MPC plans coordinated trajectories for the base and both arms at 10\,Hz over $N=5$ steps (0.75\,s horizon), with base obstacle avoidance embedded as a soft Artificial Potential Field (APF) cost term respecting the unicycle kinematics. A unified impedance--admittance controller executes these trajectories at 50\,Hz for the base and both arms, with an inter-end-effector term enforcing a minimum separation.

The main contributions of this work are:
\begin{itemize}
    \item A unified cyber-physical framework: the first to combine a VLM-RAG module, Whole-Body MPC, and unified impedance-admittance control on a bimanual mobile platform, extending the fixed-base, single-arm semantic-impedance work of SafeHumanoid~\cite{c14} and HumanoidVLM~\cite{humanoidvlm} to mobile, dual-arm operation.

    \item Semantic-to-physical parameter grounding: the VLM produces interpretable velocity, safety, and compliance parameters from scene semantics, keeping the mapping explicit rather than burying compliance in learned weights, as end-to-end models do (MoManipVLA~\cite{c16}, FALCON~\cite{c17}, HapticVLA~\cite{hapticvla}).

    \item Holistic compliance formulation: a single spring-damper law for both navigation and manipulation, with APF obstacle avoidance whose velocity and safety bounds are scaled online by the semantic layer rather than fixed at design time~\cite{c4, c5}.

    \item Experience-driven adaptation: a ChromaDB memory that retrieves validated parameter sets for similar tasks and accumulates validated experience over time, without VLM fine-tuning.
\end{itemize}

The remainder of this paper is organized as follows. Section~\ref{sec:related} reviews related work. Section~\ref{sec:system} presents the system overview. Section~\ref{sec:method} details the methodology. Section~\ref{sec:experiments} presents the experimental evaluation. Section~\ref{sec:conclusion} concludes the paper.

\section{Related Work}\label{sec:related}

\subsection{Vision-Language Models for Robot Planning}

VLMs have become powerful semantic front-ends for robotics. HYPERmotion~\cite{c1} selects robot morphology and action primitives and couples them to a motion library and MPC; VoxAct-B~\cite{c3} assigns per-arm roles in bimanual tasks from language goals; and RoboMIND~2.0~\cite{c2} trains a hierarchical VLM planner on a large dual-arm dataset. MoManipVLA~\cite{c16} and FALCON~\cite{c17} couple foundation models with Whole-Body MPC for loco-manipulation, while HapticVLA~\cite{hapticvla} learns contact-rich manipulation offline rather than sensing it at runtime. None of these systems, however, use the VLM to generate explicit, interpretable compliance parameters (stiffness, damping, or velocity limits) that shape physical interaction.

\subsection{Whole-Body MPC for Mobile Manipulation}

Whole-Body MPC has enabled agile motion on high-DoF platforms, from quadrupedal loco-manipulation~\cite{c4, c6} to task-priority transitions in mobile robots~\cite{c5}. In these frameworks, the safety margins ($d_{\text{safe}}$) and velocity envelopes are fixed; we instead adapt them online by combining LiDAR-based obstacle proximity with VLM-based semantic classification.

\subsection{Impedance and Compliant Interaction}

Since Hogan~\cite{c20}, mass-spring-damper formulations have underpinned compliant interaction, with recent work optimizing impedance for precision-compliance tradeoffs~\cite{c8}, dual-arm coordination~\cite{c10}, and reinforcement-learned variable impedance (DA-VIL~\cite{c18}). SafeHumanoid~\cite{c14} and HumanoidVLM~\cite{humanoidvlm} ground impedance scheduling in a VLM-RAG loop but remain fixed-base and single-arm. GenerativeMPC unifies the impedance-admittance formulation across the whole mobile platform, base and arms, in a single consistent approach to safe, context-aware operation.

\section{System Overview}\label{sec:system}

\subsection{Hardware Platform}

\begin{figure}[t!]
    \centering
    \includegraphics[width=0.6\columnwidth]{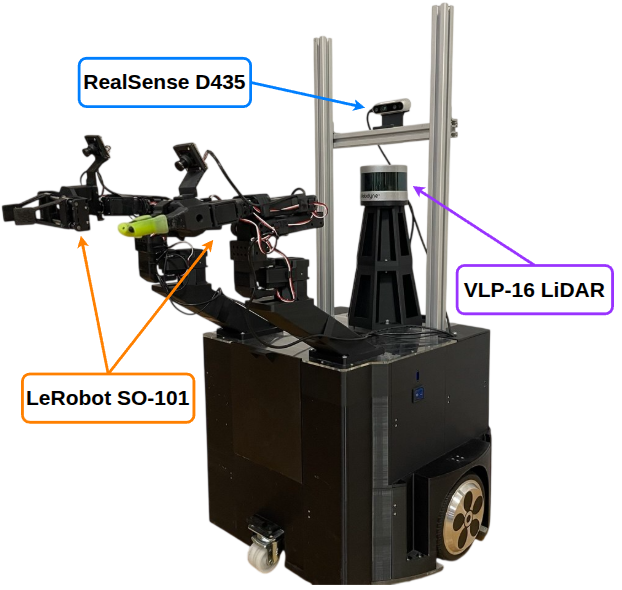}
    \caption{GenerativeMPC hardware platform: a differential-drive base with two SO-101 manipulators (five-DoF arm and one-DoF gripper, Feetech STS3215 servos), a Velodyne VLP-16 LiDAR, a RealSense D435 RGB-D camera, and an NVIDIA Jetson Orin NX.}
    \label{fig:platform}
    \vspace{- 0.5 cm} 
\end{figure}

\begin{figure}[t]
    \centering
     \includegraphics[width=0.85\columnwidth]{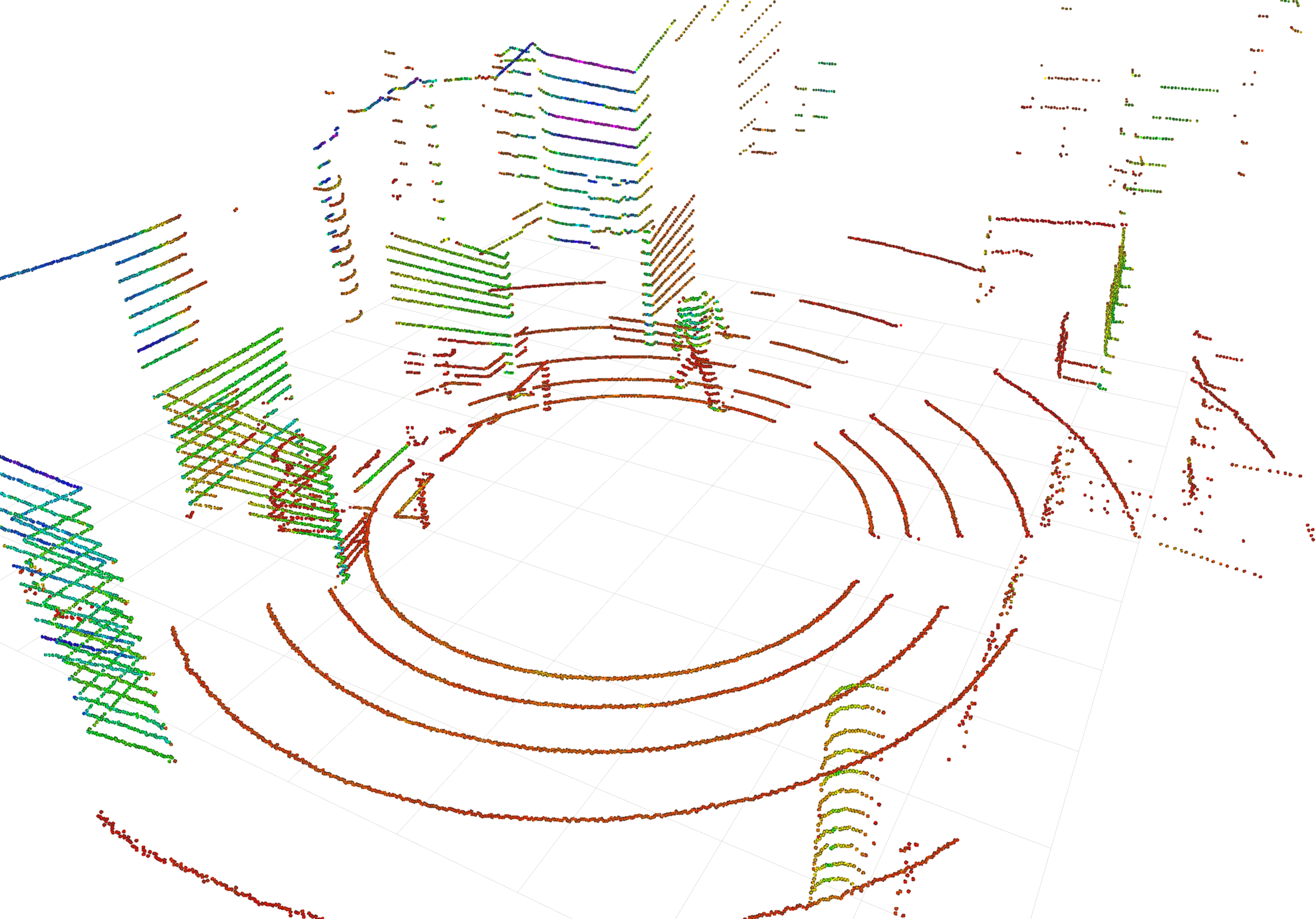}
    \caption{Velodyne VLP-16 point cloud of the workspace, used for
    obstacle detection in the MPC/APF cost.}
    \label{fig:lidar}
    \vspace{- 0.5 cm} 
\end{figure}

The platform consists of a 22\,kg differential-drive base with a wheel radius of
$r = 0.085$\,m and a wheelbase of $L = 0.455$\,m, together with two SO-101 serial manipulators~\cite{cadene2024lerobot}, each comprising a five-DoF arm and a one-DoF gripper actuated by six Feetech STS3215 servos on independent serial buses (Fig.~\ref{fig:platform}). A Velodyne VLP-16 LiDAR provides point-cloud data to the MPC/APF control layer for collision avoidance (Fig.~\ref{fig:lidar}), while a RealSense D435 RGB-D camera provides visual input to the VLM layer for semantic understanding. All computation runs onboard an NVIDIA Jetson Orin NX (16\,GB) under ROS\,2 Humble.

\subsection{Three-Layer Architecture}

GenerativeMPC organizes control into three hierarchical layers illustrated in Fig.~\ref{fig:architecture}: VLM-RAG for semantic parameter generation, Whole-Body MPC for trajectory optimization at 10\,Hz over $N=5$ steps, and Impedance-Admittance for compliant tracking at 50\,Hz. Completed episodes are stored in ChromaDB and retrieved to inform future VLM queries.

\subsection{Simulation and Sim-to-Real}

Robot models were built in SolidWorks and exported to the Unified Robot Description Format (URDF), then to MuJoCo XML (MJCF) and Isaac Sim Universal Scene Description (USD) with identical geometry, inertia, and joint limits. MuJoCo was used for controller development and Isaac Sim for full-pipeline validation of the complete architecture. In simulation, the MPC, impedance controller, and physics engine run in a single 50\,Hz callback; on hardware, these components are distributed across multiple ROS\,2 nodes, where the 30--60\,ms Modbus RTU latency acts as a time delay in the impedance loop and destabilizes gains that are stable in simulation, so all impedance parameters were re-tuned for stable operation (Table~\ref{tab:simreal_gains}).

\begin{table}[!htb]
\caption{Sim-to-Real Gain Comparison}
\label{tab:simreal_gains}
\centering\small
\setlength{\tabcolsep}{1.5pt}
\begin{tabular}{|l|c|@{\hspace{1pt}}c@{\hspace{1pt}}|}
\hline
\textbf{Parameter} & \textbf{Simulation} & \textbf{Real} \\
\hline
\multicolumn{3}{|l|}{\textit{Base --- navigation (sim and real)}} \\
\hline
$K_i,\,D_i$            & 5.0, 25.0   & 25.0, 35.0 \\
$K_a,\,D_a$            & 15.5, 35.0  & 15.5, 25.0 \\
$k_\theta$             & 7.5         & 3.5  \\
\hline
\multicolumn{3}{|l|}{\textit{Arms --- manipulation (sim and proposed hardware gains)}} \\
\hline
$K_i,\,D_i$            & 12.0, 65.0  & 1.5, 25.0 \\
$K_a,\,D_a$            & 0.12, 35.0  & 0.3, 40.0 \\
$v_{\text{EE}},\,d_{\text{self}},\,\beta$ & 0.025, 0.08, 0.5 & 0.02, 0.08, 0.5 \\
\hline
\multicolumn{3}{l}{$k_\theta$: heading gain; $v_{\text{EE}}$: maximum end-effector speed (m/s);} \\
\multicolumn{3}{l}{$d_{\text{self}}$: minimum inter-end-effector distance (m); $\beta$: separation gain.} \\
\end{tabular}
\end{table}

\section{Methodology}\label{sec:method}

\subsection{Layer 1: VLM--RAG Task Planning}

\textbf{Scene interpretation.} An edge-deployed InternVL3-1B model (INT4-quantized) maps the instruction and RGB frame to a structured JSON
description: task type, object class, a coarse fragility label (from an object registry), a human-presence flag, and an inference confidence score. The VLM does not estimate human \emph{distance}; this is measured from the LiDAR cloud and enters Layer~2 through the obstacle set $\mathcal{O}(t)$ in the APF cost~(\ref{eq:apf}).

\textbf{Retrieval.} The JSON is embedded with all-MiniLM-L6-v2 ($384$-d) and matched by cosine similarity against a ChromaDB store of validated tuples $\Pi=(v_{\max},\omega_{\max},d_{\text{safe}},K_i,D_i)$; the motion limits route to the MPC envelope and the gains to the impedance--admittance controller. The parameter store is built offline; completed episodes are logged at runtime to an experience store used in later retrievals, without VLM
fine-tuning. The retrieved $d_{\text{safe}}$ represents the required clearance from the robot boundary. Before entering the APF cost, it is converted to the base-center-referenced influence radius as $\rho_{\text{0}}= d_{\text{front}} + d_{\text{safe}}$, where $d_{\text{front}}$ is the robot’s maximum forward extent from the base center.

\textbf{Bounding and fallback.} Retrieved values are validated by construction, and three mechanisms
ensure robustness. Each value is clamped to an admissible range: $d_{\text{safe}}$ is bounded below by its ISO-derived minimum, whereas $v_{\max}$ and $\omega_{\max}$ are capped at platform limits. Per-update parameter changes are rate-limited, bounding the energy variation term $\tfrac12\|\Delta K_i\|\,\|\tilde{\mathbf{x}}\|^2$ discussed in Sec.~\ref{subsec:stability}. If the cosine distance of the best match exceeds the threshold, the layer retains the last validated set. Updates are triggered only on task transitions, and in Layer~2 the limits act purely as constraint bounds, so the optimizer cannot be driven infeasible.


\textbf{Human-aware policy.} Awareness is local, not global: the MPC bounds stay nominal, but within $1.5$\,m of a detected human, a speed governor caps the command at $0.10$\,m/s, the human profile stiffens base tracking ($K_i\!:\!5\!\to\!8$, $D_i\!:\!25\!\to\!35$), and the base-center-referenced APF influence radius is selected according to the obstacle motion state: $\rho_0\!=\!0.78$\,m for a stationary person or obstacle and $\rho_0\!=\!1.03$\,m for a moving person. A LiDAR obstacle within $1.5$\,m triggers a VLM classification pass; a human or unknown object engages the human-aware profile, whereas a confirmed static object retains the nominal profile. Instruction cues such as ``carefully'' also engage it. 

\subsection{Layer 2: Whole-Body MPC}

\textbf{State and control.} The state $\mathbf{x}\in\mathbb{R}^{9}$ stacks the base pose and both end-effector (EE) positions, while the control $\mathbf{u}\in\mathbb{R}^{8}$ stacks the base and EE velocities:
\begin{equation}
\mathbf{x}=[x_b,y_b,\theta,\mathbf{p}_L^\top,\mathbf{p}_R^\top]^\top,\quad
\mathbf{u}=[v,\omega,\dot{\mathbf{p}}_L^\top,\dot{\mathbf{p}}_R^\top]^\top .
\end{equation}
The base follows a unicycle model and each EE a first-order integrator, over $N=5$ steps at $\Delta t=0.15$\,s ($0.75$\,s horizon).

\textbf{Cost.}
$J=J_{\text{term}}+J_{\text{run}}+J_{\text{obs}}+J_{\text{ctrl}}+J_{\text{smooth}}$.
The terminal and running terms are weighted quadratics in base-pose, EE, and heading error ($w_p=5000,\ w_\theta=25,\ w_{\text{arm}}=200,\ w_{\text{path}}=8$); the control and smoothness terms penalize effort and step-to-step change. Obstacle avoidance is a soft Khatib APF~\cite{c13} over $\mathcal{O}(t)$:
\begin{equation}\label{eq:apf}
U_{\text{rep}}(\mathbf{x}_k)=
\begin{cases}
\tfrac{\eta}{2}\!\left(\tfrac{1}{\rho(\mathbf{x}_k)}-\tfrac{1}{\rho_0}\right)^{2},
& \rho(\mathbf{x}_k)\le\rho_0\\[4pt]
0,&\text{otherwise,}
\end{cases}
\end{equation}
with $\eta=50$ fixed and $\rho_0$ context-selected by Layer~1. As the nonholonomic constraint projects out lateral repulsion, avoidance is planned in the cost, not applied reactively. The problem is solved at 10\,Hz by Sequential Least Squares Programming (SLSQP), warm-started from the previous solution; the limits enter as hard constraints, so an aggressive value is projected onto the feasible set and a constraint-violating warm-start is discarded.

\textbf{Safety-clearance derivation.} The minimum robot-boundary clearance $d_{\text{safe}}$ is computed from the protective-separation terms in ISO/TS\,15066~\cite{iso_ts15066} and ISO\,13855~\cite{iso13855} as $d_{\text{safe}} = S_h+S_r+S_s+S_d$, where $S_h=v_{h}t_{lat}$ and $S_r=v_{r}t_{lat}$ represent the human and robot travel during the system latency, $S_s = v_r^2/(2a_{\max})$ is the robot braking distance, and $S_d$ is the sensing uncertainty margin. Using $v_h=1.6$\,m/s, $v_r=0.10$\,m/s, $t_{\text{lat}}=0.16$\,s, $a_{\max}=0.5$\,m/s$^2$, $S_d=0.10$\,m, and $d_{front}=0.65$\,m gives a moving-person clearance of $d_{safe}^{\text{moving}} = 0.382$\,m and a corresponding base-center-referenced APF influence radius of $\rho_0^{\text{moving}}=1.032\approx1.03$\,m. For a stationary person or obstacle ($v_h=0$), the clearance is $d_{safe}^{\text{stationary}} = 0.126$\,m and the corresponding influence radius is $\rho_0^{\text{stationary}}=0.776\approx0.78$\,m. The ISO\,3691-4~\cite{iso3691} racking clearance case uses a radius of $1.15$\,m. Layer 1 selects the appropriate radius according to the detected obstacle class and motion state.


\subsection{Layer 3: Unified Impedance--Admittance Control}

A single virtual spring--damper governs the base and both arms. From the MPC reference $(\mathbf{x}_d,\dot{\mathbf{x}}_d)$ and measured state, \begin{equation}
\begin{aligned}
\mathbf{F}_{\text{virt}} &= D_i(\dot{\mathbf{x}}_d-\dot{\mathbf{x}})
+K_i(\mathbf{x}_d-\mathbf{x}), \\
\dot{\mathbf{x}}_c &= \dot{\mathbf{x}}_d
+D_a^{-1}\!\big(\mathbf{F}_{\text{virt}}-K_a(\mathbf{x}_c-\mathbf{x}_d)\big).
\end{aligned}
\end{equation}

The commanded velocity is integrated at 50\,Hz with $(K_i,D_i)$ from Layer~1 and no force sensors. For the base, $\dot{\mathbf{x}}_c$ is projected onto the heading as $v_c=\cos\theta\,\dot{x}_c +\sin\theta\,\dot{y}_c $, with $\omega_c=k_\theta\,\mathrm{wrap}(\theta_d-\theta)$ and differential-drive wheel kinematics.

\textbf{Arm IK and inter-end-effector separation.} Each arm is represented by a reduced Pinocchio model with the non-arm joints locked, while the gripper is commanded separately. The corrected EE reference is reached using a damped least-squares (Levenberg--Marquardt) solve on a stacked, weighted position--orientation residual,
\begin{equation}
\begin{gathered}
\Delta\mathbf{q}=\tilde{\mathbf{J}}^{\top}
\big(\tilde{\mathbf{J}}\tilde{\mathbf{J}}^{\top}+\lambda\mathbf{I}\big)^{-1}\tilde{\mathbf{e}},
\\[4pt]
\tilde{\mathbf{J}}=\begin{bmatrix}\alpha_p\mathbf{J}_p\\ \alpha_R\mathbf{J}_{R,\mathcal{A}}\end{bmatrix},
\qquad
\tilde{\mathbf{e}}=\begin{bmatrix}\alpha_p\mathbf{e}_p\\ \alpha_R\mathbf{e}_{R,\mathcal{A}}\end{bmatrix},
\end{gathered}
\end{equation}
where $\mathbf{e}_p=\mathbf{p}^{\text{des}}-\mathbf{p}(\mathbf{q})$, $\mathbf{e}_{R,\mathcal{A}}=\mathrm{Log}(\mathbf{R}^{\text{des}}\mathbf{R}(\mathbf{q})^{\top})$ on selected axes $\mathcal{A}$, and $\lambda$ damps motion near singularities. The five arm joints are redundant for the 3-D position task; the low-weight orientation residual ($\alpha_p=1.0\gg \alpha_R=0.04$) biases the solution toward the nominal tool orientation without enforcing exact tracking. The damped minimum-norm update penalizes excessive joint motion, with updates integrated on the joint manifold and clamped to the joint limits. The two arms are solved independently and coordinated at the task level by the MPC. Before each IK call, if the two EE references are closer than  $d_{\text{self}}=0.08$\,m they are separated symmetrically by $\beta\delta$ along their connecting axis ($\beta=0.5$, penetration depth $\delta$), embedding the inter-end-effector separation constraint at the reference level.

\textbf{Stability under parameter switching.}\label{subsec:stability} The gains $(K_i,D_i)$ are piecewise constant, updated only on task transitions and never stepped between cycles. Within each interval the law is passive~\cite{c20}, with storage function $V=\tfrac12\tilde{\mathbf{x}}^\top K_i\tilde{\mathbf{x}}$ ($\tilde{\mathbf{x}}=\mathbf{x}_d-\mathbf{x}$) dissipated by $D_i\!\succ\!0$. Since $\mathbf{x}_c$ stays continuous across a switch, no impulsive command is issued and the injected energy is bounded by $\tfrac12\|\Delta K_i\|\,\|\tilde{\mathbf{x}}\|^2$, kept finite by the rate-limited gain change and dissipated by the damping. Energy injection is thus bounded at every switch and passivity holds within each interval.

\begin{table}[!ht]
\centering
\caption{Whole-Body MPC Parameters}\label{tab:mpc_params}
\setlength{\tabcolsep}{4pt}\small
\begin{tabular}{|l|c|l|}
\hline
\textbf{Parameter} & \textbf{Value} & \textbf{Description}\\
\hline
$N$ & 5 & Prediction horizon\\
$\Delta t$ & 0.15\,s & Time step\\
$v_{\max},\omega_{\max}$ & 0.25\,m/s, 1.5\,rad/s & Base velocity limits\\
$\dot p_{\max}$ & 0.012\,m/s & Maximum EE velocity\\
$w_p,w_\theta,w_{\text{arm}},w_{\text{path}}$ & 5000, 25, 200, 8 & Cost weights\\
$\eta$ & 50 & APF gain (fixed)\\
$\rho_0$ & 0.78 / 1.03 / 1.15\,m & APF radius \\
\hline
\end{tabular}
\vspace{-6mm}
\end{table}

\section{Experimental Evaluation}\label{sec:experiments}


Experiments are conducted in Isaac Sim and on the physical platform; MuJoCo is used only for controller development (Sec.~\ref{sec:system}), and ``simulation'' below refers to Isaac Sim. The VLM-RAG module is seeded with five demonstration episodes (free-space navigation, human-proximate navigation, and bimanual pick-and-place), forming the initial ChromaDB store. Navigation metrics are reported over $n=5$ trials per condition, and manipulation metrics over $n=10$ trials per arm. Tolerances are $20$\,mm (position), $2.0$ deg. (heading), and $5$\,mm (end-effector).

\subsection{Computational Performance}

Table~\ref{tab:compute} reports component latency on the Jetson Orin NX; each component's mean latency stays below its execution budget. The $3.5$\,s VLM inference is acceptable because the module is asynchronous,
triggered only on task transitions rather than at the control rate.

\begin{table}[!htb]
\caption{System Computational Performance (Jetson Orin NX)}
\label{tab:compute}
\centering\small
\begin{tabular}{|l|c|c|}
\hline
\textbf{Component} & \textbf{Mean $\pm$ SD [ms]} & \textbf{Budget [ms]}\\
\hline
MPC solve        & $75.0 \pm 30.0$   & 100 \\
LiDAR processing & $15.0 \pm 8.0$    & 100 \\
VLM inference    & $3500 \pm 650$    & 5000 \\
\hline
\end{tabular}
\vspace{- 0.3 cm} 
\end{table}

\begin{figure}[!t]
    \centering
    \includegraphics[width=0.8\columnwidth]{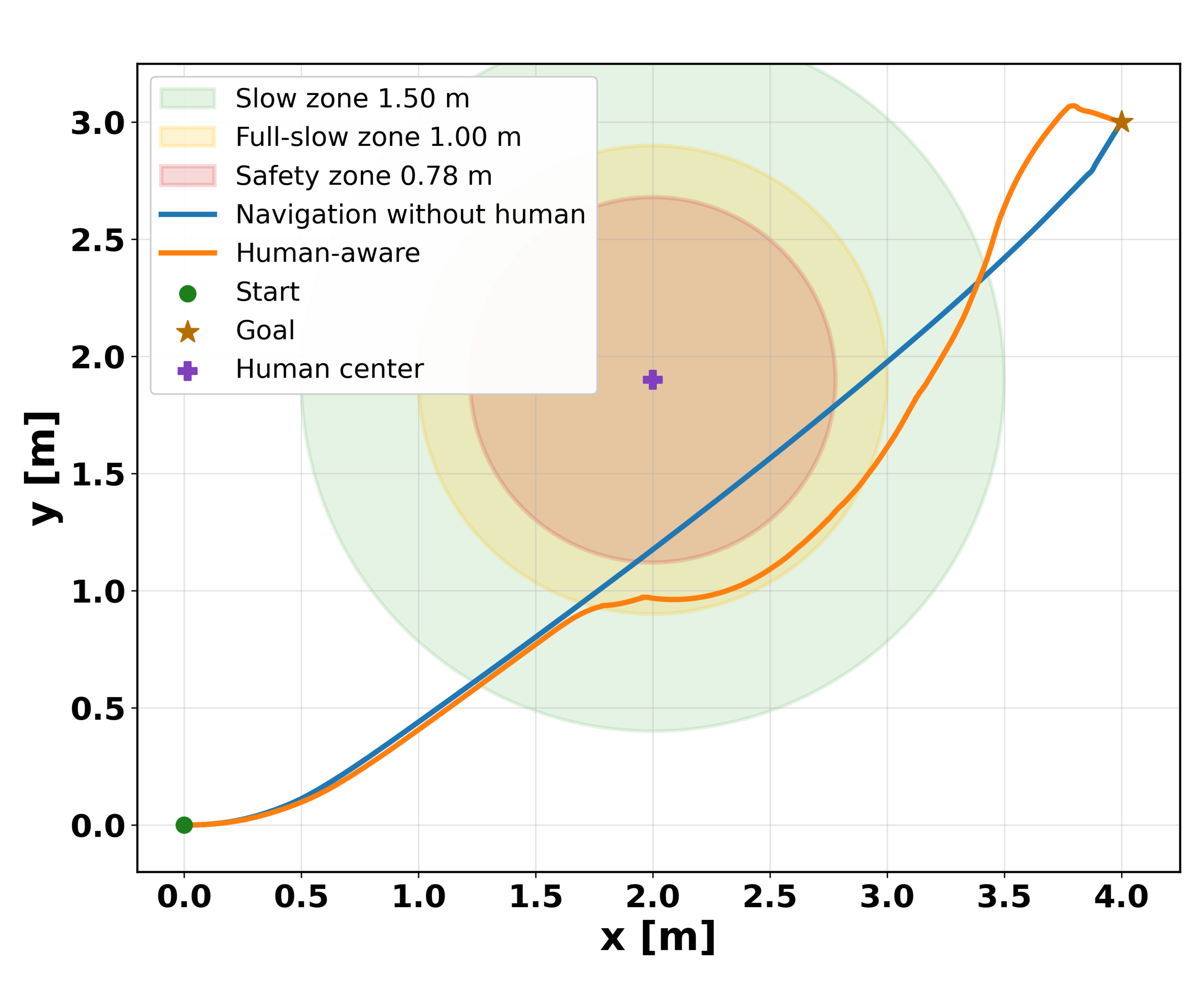}
    \caption{Human-aware navigation in Isaac Sim: base trajectories to the goal $(4.0,\,3.0)$\,m for free-space (blue) and human-aware (orange) navigation. Zones around the stationary person mark the slow (1.50\,m), full-slow (1.00\,m), and stationary-case APF threshold (0.78\,m); the human-aware path steers around the person via the APF cost.}
    \label{fig:traj}\
\vspace{- 0.5 cm}    
\end{figure}

\subsection{Free-Space and Human-Aware Navigation}

The robot navigates from $(0,0)$ to $(4.0,3.0)$\,m under two conditions (Fig.~\ref{fig:traj}). Free navigation plans a direct path at the nominal cap $v_{\max}=0.25$\,m/s. In the human-aware condition, LiDAR detects a nearby obstacle, and the VLM-RAG module classifies it as a human and retrieves a conservative profile, reducing the velocity cap to $0.10$\,m/s (a $60\%$ reduction), while the APF cost deflects the base around the selected safety zones. Both conditions converge within tolerance: position errors are $14.09\pm0.90$\,mm (free) and
$12.05\pm1.83$\,mm (human-aware). The slightly lower error in the human-aware condition may result from the smoother, lower-speed approach.

\textbf{Safety and the cost of adaptation.} Across all stationary-person trials, the minimum base-center-to-person distance was $0.879\pm0.041$\,m, exceeding the stationary-case base-center-referenced APF threshold of $0.776$\,m in every trial. The free and human-aware conditions also isolate the effect of the semantic safety scaling relative to an MPC with static nominal bounds: the nominal profile reaches the goal in $31.99\pm1.62$\,s, whereas the adaptive profile takes $46.36\pm2.59$\,s---approximately $45\%$ more time in exchange for maintaining the selected stationary-case threshold.

\subsection{Sim-to-Real Navigation Transfer}

The same navigation experiments were repeated on the physical platform ($n=5$ each). Position errors increased to $17.0\pm1.4$\,mm (free) and $17.8\pm2.0$\,mm (human-aware), remaining within the $20$\,mm tolerance;
the degradation is consistent with the $30$--$60$\,ms Modbus RTU latency
(Sec.~\ref{sec:system}). The real platform reproduces the $60\%$ velocity reduction while
maintaining the $0.776$\,m stationary-case threshold in the stationary-person trials, and a separate moving-person trial shows the human-aware behavior remains feasible with a dynamic obstacle.

\subsection{Bimanual Manipulation in Simulation}

Following base convergence, the robot performs a bimanual
pick-and-place task in Isaac Sim (Fig.~\ref{fig:teaser}). The
VLM-RAG module retrieves object-class-dependent impedance gains.
Across $n=10$ trials per arm, the final placement errors were
$3.75\pm0.22$\,mm for the left arm and $3.85\pm0.14$\,mm for
the right arm. The system achieved pick, place, and overall
task-success rates of $90\%$, $85\%$, and $85\%$, respectively,
while maintaining the specified $0.08$\,m inter-end-effector
clearance.

\subsection{Comparison with Related Mobile-Manipulation Frameworks}


The principal distinction between GenerativeMPC and end-to-end mobile-manipulation frameworks such as MoManipVLA~\cite{c16} is architectural. GenerativeMPC represents semantic decisions as explicit
velocity limits, safety margins, and impedance parameters, which are subsequently enforced through constrained MPC and compliant control. A direct numerical comparison is not reported because the tasks, robotic platforms, and evaluation protocols differ.

\begin{table}[!htb]
\caption{Consolidated Performance Summary}
\label{tab:results}
\centering\small
\setlength{\tabcolsep}{3pt}
\begin{tabular}{|l|c|c|}
\hline
\textbf{Metric} & \textbf{Sim} & \textbf{Real} \\
\hline
\multicolumn{3}{|l|}{\textit{Navigation ($N=5$ per condition)}} \\
\hline
Pos.\ error --- free [mm]        & $14.09\pm0.90$ & $17.0\pm1.4$ \\
Pos.\ error --- human-aware [mm] & $12.05\pm1.83$ & $17.8\pm2.0$ \\
Heading error [deg]              & $<2.0$ & $<2.0$ \\
Velocity reduction               & $60\%$ & $60\%$ \\
Min.\ center-person dist.\ [m]   & $0.879\pm0.041$ & $0.855\pm0.045$ \\
APF threshold ($\geq0.776$\,m)   & \checkmark & \checkmark \\
\hline
\multicolumn{3}{|l|}{\textit{Manipulation ($N=10$ per arm, sim only)}} \\
\hline
EE error --- left [mm]   & $3.75\pm0.22$ & --- \\
EE error --- right [mm]  & $3.85\pm0.14$ & --- \\
Overall task success     & $85\%$ & --- \\
\hline
\end{tabular}
\vspace{-0.3cm}
\end{table}

\section{Conclusion}\label{sec:conclusion}

This paper has introduced GenerativeMPC, a hierarchical
cyber-physical framework that grounds semantic scene understanding
in interpretable control parameters for compliant bimanual mobile
manipulation. A VLM-RAG module derives velocity and safety constraints for the
Whole-Body MPC and stiffness and damping gains for the
impedance-admittance controller from imagery and language, without
manual scheduling or retraining, while a unified spring-damper
formulation governs base tracking and arm compliance. Human-aware navigation was validated in simulation and
on hardware, achieving errors below 20\,mm and a 60\% velocity
reduction near a detected person. In simulation, bimanual
manipulation achieved an 85\% overall task-success rate with final
end-effector errors below 5\,mm. Future work will validate
manipulation on hardware, broaden the ChromaDB task coverage, and
reduce VLM inference latency.

\section*{Acknowledgments} 
Research reported in this publication was financially supported by the RSF grant No. 24-41-02039.

\balance  
\bibliographystyle{IEEEtran} 
\bibliography{ref} 

@IEEEtranBSTCTL{IEEEexample:BSTcontrol,
  CTLuse_forced_etal       = "yes",
  CTLmax_names_forced_etal = "6",
  CTLnames_show_etal       = "1"
}

@misc{c1,
      title={{HYPER}motion: Learning Hybrid Behavior Planning for Autonomous Loco-manipulation}, 
      author={Jin Wang and Rui Dai and Weijie Wang and Luca Rossini and Francesco Ruscelli and Nikos Tsagarakis},
      year={2024},
      note={arXiv:2406.14655}}

@misc{c2,
      title={Robo{MIND} 2.0: A Multimodal, Bimanual Mobile Manipulation Dataset for Generalizable Embodied Intelligence}, 
      author={Chengkai Hou and Kun Wu and Jiaming Liu and Zhengping Che and Di Wu and Fei Liao and Guangrun Li and Jingyang He and Qiuxuan Feng and Zhao Jin and others},
      year={2025},
      note={arXiv:2512.24653}}

@misc{c3,
      title={Vox{A}ct-{B}: Voxel-Based Acting and Stabilizing Policy for Bimanual Manipulation}, 
      author={I-Chun Arthur Liu and Sicheng He and Daniel Seita and Gaurav Sukhatme},
      year={2024},
      note={arXiv:2407.04152}}

@INPROCEEDINGS{c4,
   title={{Whole-Body MPC} for Highly Redundant Legged Manipulators: Experimental Evaluation with a 37 DoF Dual-Arm Quadruped},
   DOI={10.1109/humanoids57100.2023.10375215},
   booktitle={Proc. IEEE-RAS Int. Conf. on Humanoid Robots (Humanoids)},
   author={Dadiotis, Ioannis and Laurenzi, Arturo and Tsagarakis, Nikos},
   pages={1-8},
   year={2023}
    }

@INPROCEEDINGS{c5,
  author={Wang, Yushi and Chen, Ruoqu and Zhao, Mingguo},
  booktitle={Proc. IEEE Int. Conf. on Robotics and Automation (ICRA)}, 
  title={Whole-Body Model Predictive Control for Mobile Manipulation With Task Priority Transition}, 
  year={2025},
  volume={},
  number={},
  pages={13356-13362},
  doi={10.1109/ICRA55743.2025.11127515}}

@INPROCEEDINGS{c6,
  author={Chiu, Jia-Ruei and Sleiman, Jean-Pierre and Mittal, Mayank and Farshidian, Farbod and Hutter, Marco},
  booktitle={Proc. IEEE Int. Conf. on Robotics and Automation (ICRA)}, 
  title={A Collision-Free {MPC} for Whole-Body Dynamic Locomotion and Manipulation}, 
  year={2022},
  volume={},
  number={},
  pages={4686-4693},
  doi={10.1109/ICRA46639.2022.9812280}}

@INPROCEEDINGS{c7,
  author={Zhuang, Zixuan and Zheng, Le and Li, Wanyue and Liu, Renming and Lu, Peng and Cheng, Hui},
  booktitle={Proc. IEEE Int. Conf. on Robotics and Automation (ICRA)}, 
  title={{RM-Planner}: Integrating Reinforcement Learning with Whole-Body Model Predictive Control for Mobile Manipulation}, 
  year={2025},
  volume={},
  number={},
  pages={7263-7269},
  doi={10.1109/ICRA55743.2025.11127719}}

@ARTICLE{c8,
  author={Jin, Zhehao and Qin, Dongdong and Liu, Andong and Zhang, Wenan-An and Yu, Li},
  journal={IEEE/ASME Transactions on Mechatronics}, 
  title={Model Predictive Variable Impedance Control of Manipulators for Adaptive Precision-Compliance Tradeoff}, 
  year={2023},
  volume={28},
  number={2},
  pages={1174-1186},
  doi={10.1109/TMECH.2022.3204350}}

@article{c10,
author = {Xin Jing and Loris Roveda and Jianfei Li and Yaobing Wang and Haibo Gao},
title ={An adaptive impedance control for dual-arm manipulators incorporated with the virtual decomposition control},
journal = {Journal of Vibration and Control},
volume = {30},
number = {11-12},
pages = {2647-2660},
year = {2024},
doi = {10.1177/10775463231182462}}

@inproceedings{c13,
  author={Khatib, O.},
  booktitle={Proc. IEEE Int. Conf. on Robotics and Automation (ICRA)}, 
  title={Real-time obstacle avoidance for manipulators and mobile robots}, 
  year={1985},
  volume={2},
  number={},
  pages={500-505},
  doi={10.1109/ROBOT.1985.1087247}}

@inproceedings{c14,
author = {Mahmoud, Yara and Sam, Jeffrin and Nguyen, Khang and Fernando, Marcelino Julio and Tokmurziyev, Issatay and Altamirano Cabrera, Miguel and Khan, Muhammad Haris and Lykov, Artem and Tsetserukou, Dzmitry},
title = {{SafeHumanoid: VLM-RAG-Driven} Impedance Control of Humanoid Robot},
year = {2026},
booktitle = {Proc. ACM/IEEE Int. Conf. on Human-Robot Interaction (HRI)},
pages = {974-978},
numpages = {5}}

@INPROCEEDINGS{c16,
  author    = {Wu, Zhenyu and Zhou, Yuheng and Xu, Xiuwei and Wang, Ziwei and Yan, Haibin},
  booktitle = {Proc. IEEE/CVF Conf. on Computer Vision and Pattern Recognition (CVPR)},
  title     = {{MoManipVLA}: Transferring Vision-Language-Action Models for General Mobile Manipulation},
  year      = {2025},
  pages     = {1714--1723},
  doi       = {10.1109/CVPR52734.2025.00167}
}

@misc{c17,
  title={{FALCON}: Actively Decoupled Visuomotor Policies for Loco-Manipulation with Foundation-Model-Based Coordination},
  author={He, Chengyang and Sun, Ge and Bai, Yue and Lu, Junkai and Zhao, Jiadong and Sartoretti, Guillaume},
  note={arXiv:2512.04381},
  year={2025}
}

@INPROCEEDINGS{c18,
  author={Karim, Md Faizal and Bollimuntha, Shreya and Hashmi, Mohammed Saad and Das, Autrio and Singh, Gaurav and Sridhar, Srinath and Singh, Arun Kumar and Govindan, Nagamanikandan and Krishna, K Madhava},
  booktitle={Proc. IEEE Int. Conf. on Robotics and Automation (ICRA)}, 
  title={{DA-VIL}: Adaptive Dual-Arm Manipulation with Reinforcement Learning and Variable Impedance Control}, 
  year={2025},
  volume={},
  number={},
  pages={11896-11903},
  doi={10.1109/ICRA55743.2025.11127487}}

@INPROCEEDINGS{c20,
  author={Hogan, Neville},
  booktitle={Proc. American Control Conf.}, 
  title={Impedance Control: An Approach to Manipulation}, 
  year={1984},
  volume={},
  number={},
  pages={304-313},
  doi={10.23919/ACC.1984.4788393}}

@misc{cadene2024lerobot,
    author = {Cadene, Remi and Alibert, Simon and Soare, Alexander and Gallouedec, Quentin and Zouitine, Adil and Palma, Steven and Kooijmans, Pepijn and Aractingi, Michel and Shukor, Mustafa and Aubakirova, Dana and Russi, Martino and Capuano, Francesco and Pascal, Caroline and Choghari, Jade and Moss, Jess and Wolf, Thomas},
    title = {{LeRobot}: State-of-the-art Machine Learning for Real-World Robotics in {PyTorch}},
    howpublished = "[Online]. Available: \url{https://github.com/huggingface/lerobot}"
}

@inproceedings{c_internvl,
  author={Chen, Zhe and Wu, Jiannan and Wang, Wenhai 
          and Su, Weijie and Chen, Guo and Xing, Sen 
          and Zhong, Muyan and Liu, Qinglong and Lu, 
          Lewei and Li, Bin and others},
  title={Intern{VL}: Scaling up Vision Foundation Models 
         and Aligning for Generic Visual-Linguistic Tasks},
  booktitle={Proc. IEEE/CVF Conf. on Computer Vision 
             and Pattern Recognition (CVPR)},
  year={2024},
  pages={24185--24198}}

@inproceedings{humanoidvlm,
  author    = {Mahmoud, Yara and Yaqoot, Yasheerah and Altamirano Cabrera, Miguel and Tsetserukou, Dzmitry},
  title     = {Humanoid{VLM}: Vision-Language-Guided Impedance Control for Contact-Rich Humanoid Manipulation},
  booktitle = {Companion Proc. ACM/IEEE Int. Conf. on Human-Robot Interaction (HRI)},
  year      = {2026}, 
  pages = {918--922}}

@misc{hapticvla,
  title  = {Haptic{VLA}: Contact-Rich Manipulation via Vision-Language-Action Model without Inference-Time Tactile Sensing},
  author = {Gubernatorov, Konstantin and Sannikov, Mikhail and Mikhalchuk, Ilya and Kuznetsov, Egor and Artemov, Makar and Ouwatobi, Ogunwoye Faith and Fernando, Marcelino and Asanov, Artem and Guo, Ziang and Tsetserukou, Dzmitry},
  year  = {2026}, 
  note = {arXiv:2603.15257}}

@standard{iso_ts15066,
  organization = {{International Organization for Standardization}},
  title        = {Robots and robotic devices --- Collaborative robots},
  number       = {{ISO/TS 15066:2016}},
  type         = {Technical Specification},
  year         = {2016}
}

@standard{iso13855,
  organization = {{International Organization for Standardization}},
  title        = {Safety of machinery --- Positioning of safeguards with respect to the approach speeds of parts of the human body},
  number       = {ISO 13855:2010},
  year         = {2010}
}

@standard{iso3691,
  organization = {{International Organization for Standardization}},
  title        = {Industrial trucks --- Safety requirements and verification --- Part 4: Driverless industrial trucks and their systems},
  number       = {ISO 3691-4:2020},
  year         = {2020}
}

\end{document}